\documentclass[letterpaper, 10 pt, conference]{ieeeconf}  

\usepackage[noadjust]{cite}
\IEEEoverridecommandlockouts                              
                  
\newcommand{\PREPRINTYEAR}{2024}

\newcommand{\DOI}{10.1109/RoMoCo60539.2024.10604306} 
 
\usepackage[placement=top,vshift=-2,firstpage=true]{background}
\SetBgScale{1.0}
\SetBgContents{\parbox{0.95\textwidth}{\small \begin{center} 
    The manuscript will appear in 
        2024 13th International Workshop on Robot Motion and Control (RoMoCo)
        under DOI: \DOI\end{center} \vspace{-0.2cm}  \copyright{}~\PREPRINTYEAR~IEEE.~Personal use of this material is permitted. Permission from IEEE must be obtained for all other uses, in any current or future media, including reprinting/republishing this material for advertising or promotional purposes, creating new collective works, for resale or redistribution to servers or lists, or reuse of any copyrighted component of this work in other works.} }
\SetBgColor{black}
\SetBgAngle{0}
\SetBgOpacity{1.0}

\usepackage{graphicx}
\usepackage{graphics} 
\usepackage{amsmath} 
\usepackage{amssymb}  
\usepackage{xcolor}
\usepackage{xspace}
\usepackage{subcaption} 
\usepackage{hyperref}
\usepackage{booktabs}
\usepackage{soul}

\newcommand{\vspacebeforefigure}[0]{\vspace{2mm}}
\newcommand{\vspacebeforesubcaption}[0]{\vspace{-5mm}}
\newcommand{\vspacebetweensubfigures}[0]{\vspace{2mm}}
\newcommand{\vspaceafterfigure}[0]{\vspace{-4mm}}  
\title{\LARGE \bf
A Framework for Joint Grasp and Motion Planning in Confined Spaces 
}

\author{
    Martin Rudorfer$^1$, Ji\v{r}i Hartvich$^2$, Vojt\v{e}ch Von\'{a}sek$^2$
    \thanks{
        This work was supported in part by 
            the Czech Science Foundation (GA\v{C}R) through Research Project under Grant 22-24425S, 
            by the European Union under the project Robotics and Advanced Industrial Production (reg. no. CZ.02.01.01/00/22\_008/0004590),
            by CTU grant no SGS23/177/OHK3/3T/13,
            and by CHIST-ERA under EPSRC grant no. EP/S032487/1.
        Computational resources were provided by the e-INFRA CZ project (ID:90254), supported by the Ministry of Education, Youth and Sports of the Czech Republic.
    }
    \thanks{
        $1$: M. Rudorfer is with the Department of Applied AI \& Robotics at Aston University, Birmingham, UK.
        {\tt m.rudorfer@aston.ac.uk}
    }
    \thanks{
        $2$: J. Hartvich and V. Von\'{a}sek are with the Faculty of Electrical Engineering, Czech Technical University Prague, Prague, Czech Republic.
        {\tt \{hartvjir,vonasvoj\}@fel.cvut.cz}
    }
}

\def\q{q}

\def\qrand{\q_{rand}}
\def\qstart{\q_{start}}

\def\qgoal{\q_{goal}}

\def\C{\mathcal{C}}
\def\CF{\mathcal{C}_{free}}

\def\G{\mathcal{G}}
\def\gb{p_{goal}}
\def\gdist{d_{goal}}
\def\G{\mathcal{G}}
\def\GIK{\mathcal{G}_{ik}}
\def\JRRT{J${}^{+}$-RRT}
\def\IKRRT{IK-RRT}

\begin{document}
\maketitle
\thispagestyle{empty}
\pagestyle{empty}

\begin{abstract}

Robotic grasping is a fundamental skill across all domains of robot applications.
There is a large body of research for grasping objects in table-top scenarios, where finding suitable grasps is the main challenge.
In this work, we are interested in scenarios where the objects are in confined spaces and hence particularly difficult to reach.
Planning how the robot approaches the object becomes a major part of the challenge, giving rise to methods for joint grasp and motion planning.
The framework proposed in this paper provides 20~benchmark scenarios with systematically increasing difficulty, realistic objects with precomputed grasp annotations, and tools to create and share more scenarios.
We further provide two baseline planners and evaluate them on the scenarios, demonstrating that the proposed difficulty levels indeed offer a meaningful progression.
We invite the research community to build upon this framework by making all components publicly available as open source.

\end{abstract}

\section{Introduction}
Especially during the last decade, research has pushed the boundaries on robot capabilities to grasp objects.
A multitude of methods have been suggested for picking isolated objects~\cite{Mahler2016, Mousavian2019, Alliegro_L2G_2022},
as well as grasping objects from clutter~\cite{Fang2020, Sundermeyer2021}.
These works consider table-top scenarios where the objects are easy to reach.
When picking from bins~\cite{Cordeiro2022} or shelves~\cite{Leitner2017}, the reachability of the objects needs to be considered and a suitable approach path for the robot must be found.
We push this even further and tackle scenarios in which the objects are very difficult to reach and hence approaching the object becomes a major part of the challenge.
We aim to grasp objects in confined spaces and heavily obstructed environments; for example 
    a domestic cleaning robot retrieving a toy from under the sofa, 
    or a robot loading and unloading an industrial machine tool in a confined space.

\begin{figure}[tb]
	\centering
    \vspacebeforefigure
    \begin{subfigure}[t]{0.48\columnwidth}
       \centering
       \includegraphics[width=\textwidth]{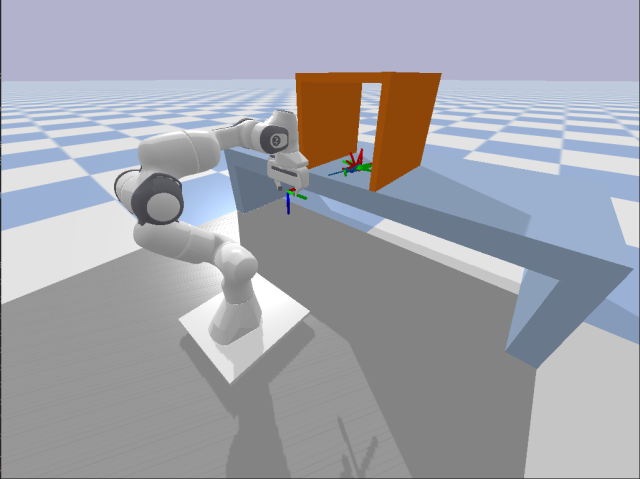}
       \vspacebeforesubcaption
       \caption{Environment 01}
  	\end{subfigure}
  	\begin{subfigure}[t]{0.48\columnwidth}
       \centering
       \includegraphics[width=\textwidth]{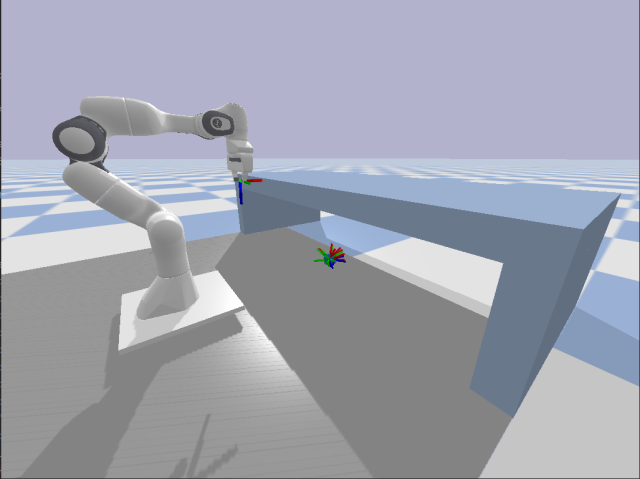}
       \vspacebeforesubcaption
       \caption{Environment 02}
  	\end{subfigure}
  	
  	\vspacebetweensubfigures
  	\begin{subfigure}[t]{0.48\columnwidth}
       \centering
       \includegraphics[width=\textwidth]{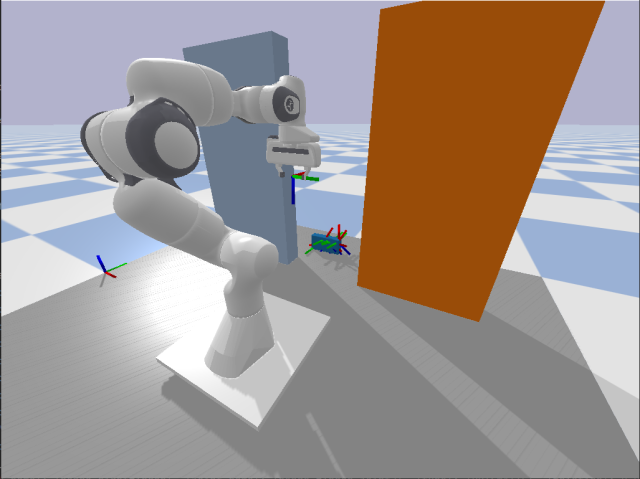}
       \vspacebeforesubcaption
       \caption{Environment 03}
  	\end{subfigure}
  	\begin{subfigure}[t]{0.48\columnwidth}
       \centering
       \includegraphics[width=\textwidth]{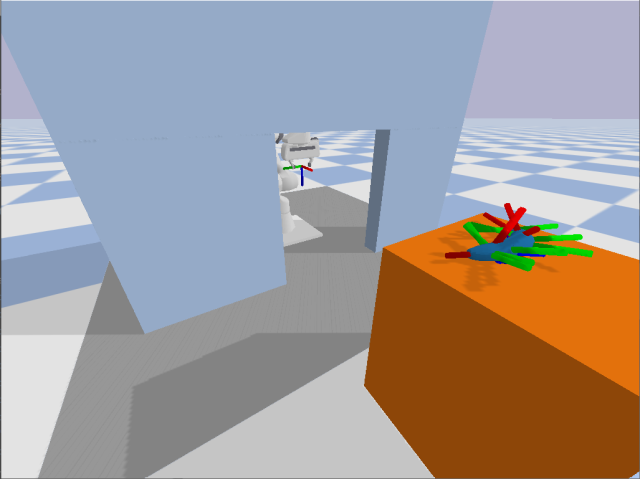}
       \vspacebeforesubcaption
       \caption{Environment 04}
  	\end{subfigure}
  	\caption{Four different environments:
        (a) grasping a screwdriver on a shelf,
        (b) grasping a ball underneath a table,
        (c) grasping a box through a narrow gap,
        (d) navigating through a narrow opening to grasp the banana.
        Each target object has a set of possible grasp candidates.
        The arm is mounted on a mobile base that can move within the grey area.
        }
  	\vspaceafterfigure
  	\label{fig:scenarios}
\end{figure}

The key to solving such challenges is an integrated approach for grasp and motion planning.
However, although very related, grasp and motion planning are mostly separate research fields.
There are a few works at the intersection~\cite{Vahrenkamp2010,huhConstrainedSamplingBasedPlanning2018,ichnowskiGOMPGraspOptimizedMotion2020}, but these works consider different types of robots and evaluate their methods on different types of scenarios.
Often, algorithms and scenarios are not made available and are only described in the publications.
Experiment replication and hence comparison between approaches is difficult, and building upon other people's work is cumbersome.
This is a known struggle in robotics research and potentially undermines scientific progress~\cite{bonsignorio2015toward}.
Our work directly fosters reproducibility by providing benchmark scenarios, baseline planners, and the tools to build own scenarios and planners.

In this paper, we provide an open-source framework for grasping in confined environments that bridges the gap between grasping and motion planning and fosters reproducibility in robotics research.
Our framework contains:
\begin{itemize}
    \item 4 different environments (see Fig.~\ref{fig:scenarios}) with 5 difficulty levels each, giving 20 benchmark scenarios in total,
    \item 200 grasp candidates with scores for each scenario,
    \item a set of baseline planners for which we evaluate performance on our scenarios,
    \item a set of tools that allow to evaluate the planners using relevant metrics, simulate the grasps, visualize results, and create and share new scenarios and grasp candidates.
\end{itemize}

All of the above are publicly available. Please visit our project page at \href{https://mrudorfer.github.io/jogramop-framework}{mrudorfer.github.io/jogramop-framework}.\footnote{
The project page may still be under construction while the submission is being reviewed.
It will be finalized upon acceptance.
}

The remainder of this paper is organized as follows.
In Section~\ref{sec:related-works} we review the state of the art in grasping, motion planning, as well as combined methods and existing benchmarks and datasets.
In Section~\ref{sec:problem-formulation} we describe the problem more formally and then present our framework in Section~\ref{sec:framework}.
The performance of the baseline planners is analysed and discussed in Section~\ref{sec:experiments} and Section~\ref{sec:conclusions} concludes the paper.

\section{Related Works}
\label{sec:related-works}
We first summarize the most important trends in grasp and motion planning, respectively.
Then, we review existing works on joint grasp and motion planning and, finally, discuss related benchmarks and datasets.

\subsection{Grasp Planning}

A grasp for a two-finger parallel gripper is typically described as the pose of the gripper in which it closes its fingers to grasp the object.
For a given object, there may be infinitely many such poses and there is no established representation to describe the entire distribution of grasps.
Therefore, almost all methods resort to finding a set of discrete candidate poses~\cite{Kleeberger2020}.
Earlier works restrict the search space to top-down grasps~\cite{yun2011efficient}, but contemporary methods typically find 6-DoF grasps~\cite{Mousavian2019, Alliegro_L2G_2022, Fang2020, Sundermeyer2021}.
This is particularly important when grasping objects from clutter or constricted spaces.

Most methods use learning-based approaches that predict a set of grasp poses from a partial observation of an object, usually point clouds from depth images~\cite{Alliegro_L2G_2022,Fang2020,Sundermeyer2021}.
They are trained with grasps that were sampled for known objects using specific grasp sampling schemes~\cite{Eppner2019_billion_grasps} and annotated either by analytic metrics such as force-closure~\cite{Nguyen1988}, or by determining grasp success in simulation~\cite{Eppner2021}.

The set of grasps obtained from state-of-the-art grasp planners can be used as starting point for joint grasp and motion planners.
However, it is important to remember that such a set is only a sample from the whole distribution of grasps.

\subsection{Motion Planning}

The most common methods for motion planning are either optimization-based or sampling-based.
In the former case, the motion planning task is formulated as an optimization problem (usually using Quadratic programming and its variants)~\cite{fusco2018constrained}.
Optimization-based approaches ensure optimal solutions, but they 
are more difficult to adapt to non-convex obstacles and additional (e.g., end-effector) constraints.

Sampling-based planners tackle the problem by randomized sampling of the high-dimensional configuration space of the robot.
Tree-based planners like Rapidly-exploring Random Tree (RRT)~\cite{Kuffner2000} and its 
variants~\cite{mcmahonSamplingbasedMotionPlanning2018a, kingstonSamplingBasedMethodsMotion2018} are popular choices.
RRT has been widely used for robotic manipulators~\cite{vandewegheRandomizedPathPlanning2007,huhConstrainedSamplingBasedPlanning2018,shkolnikPathPlanning10002009} as they allow to consider differential and task constraints.
Although the search tree is built in the configuration space, corresponding task-space positions can be computed via forward kinematics.

Planners from both categories have been applied to the problem of joint grasp and motion planning.

\subsection{Integrated Grasp and Motion Planning}

Several works follow a multi-stage paradigm, in which a grasp set is provided by an upstream grasp planner and a trajectory is found towards one of these grasps.
A naive strategy is to find a target configuration for a selected grasp using Inverse Kinematics (IK) and employ a classic motion planner to find a path.
However, this is needlessly limiting, as for a redundant robot there are many potential configurations that can reach the task-space target.
Instead, authors of~\cite{vahrenkamp2009humanoid} utilise the pseudo-inverse of the Jacobian to extend the tree towards one of multiple task-space targets.
To further relax the constraint, Berenson et al.~\cite{Berenson2011} use task space regions around the grasps as targets.
GOMP~\cite{ichnowskiGOMPGraspOptimizedMotion2020} utilizes DexNet~\cite{Mahler2019} for generating grasp candidates and optimizes trajectories to all of them in parallel using Sequential Quadratic Programming for a bin picking scenario.
In the OMG planner~\cite{Wang2020a}, the selection of the best possible grasp is integrated into the optimization of the trajectory itself.

On the other hand, single-stage methods do not rely on a pre-determined grasp set as they jointly search for feasible paths and suitable grasps.
In~\cite{Vahrenkamp2010}, they use RRT to approach the object from all directions and once a close-enough configuration is found, a grasp score is determined.
Huh et al.~\cite{huhConstrainedSamplingBasedPlanning2018} deal with obstacles by determining the side of the object with the most free space; they then use a sampling-based planner to guide the search towards this area.
The object is approached perpendicular to its major axis and a grasp is executed once the gripper is close enough.
In NGDF~\cite{Weng2023}, a neural field is learned to predict the distance to the nearest valid grasp.
This is then used as cost-term in an optimization-based planner.

Our framework accommodates both multi-stage and single-stage methods, as we provide pre-computed grasp sets as well as high-quality object models to plan grasps online.

\subsection{Benchmarks and Datasets}

As robots are being applied to a growing range of tasks, there is a constant need for datasets and benchmarks.
Datasets mainly provide training data for learning-based approaches, while benchmarks are specifically designed to evaluate and compare performance.
Amazon's picking challenge~\cite{Leitner2017} and Ocado's benchmark for fruit and vegetable picking~\cite{Mnyusiwalla2020} both offer a real set of objects to test performance of the whole robotic system, which includes both grasp and motion planning.
In contrast to these system-level benchmarks, we are more interested in comparing the performance of different algorithms, while using the same type of robot and gripper.
To maximize availability of our benchmark, we provide simulated scenarios.

There are several datasets and benchmarks for robotic grasping.
For example, the Cornell grasping dataset~\cite{yun2011efficient} contains images of objects annotated with oriented rectangles to indicate good top-down grasp candidates.
The more recent ACRONYM~\cite{Eppner2021} dataset contains a large number of simulated scenes with 6-DoF grasps that were evaluated in simulation.
GraspNet~\cite{Fang2020} provides almost 100k real RGB-D images of 190~cluttered scenes with annotated 6-DoF grasps.
Our framework does not aim to provide comprehensive training data for grasping, but we do employ the same state-of-the-art methods to generate the grasp annotations we provide along with our scenarios.

The OMPL benchmark~\cite{sucan2012the-open-motion-planning-library} contains many sampling-based motion planners and tools for comparing them.
The MotionBenchMaker~\cite{Chamzas2022} is a set of tools that allows to create own scenarios and evaluate the planners from OMPL.
These tools have been employed recently by~\cite{Yang2023} to benchmark motion planners with scenarios similar to our confined spaces.
We are inspired by this line of work and aim to offer a similar framework for joint grasp and motion planning.

\section{Problem Formulation}
\label{sec:problem-formulation}

\begin{figure}[thpb]
  \centering
  \includegraphics[width=0.6 \columnwidth]{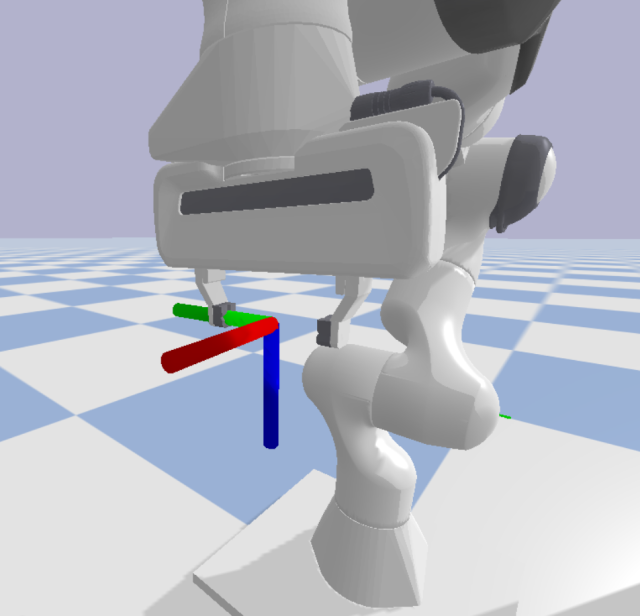}
  \caption{Visualization of the gripper frame. The y-axis (green) represents the grasp axis along which the fingers close, and the z-axis (blue) represents the gripper's approach direction.
}
  \label{fig:gripper-frame}
\end{figure}

The specific problem we consider in this work is for a mobile manipulator arm with a two-finger parallel gripper to grasp a known target object in a fully-known environment.
We assume availability of a set $\G$ of discrete grasps, where each grasp $g=(T_g, s_g)$.
The grasp pose $T_g \in SE(3)$ corresponds to the gripper frame as depicted in Fig.~\ref{fig:gripper-frame}.
The score~$s_g$ is optional and estimates the grasp stability.
This definition ensures compatibility with a large body of state-of-the-art 6-DoF grasp planners from which the set~$\G$ can be obtained.

Motion planning  is solved by searching the configuration (joint) space.
Let $\C$ denote the configuration space, where each configuration $\q \in \C$ is the set of robot joints, let
$\CF \subseteq \C$ denote the collision-free region. 
The task of the planners is to find a path from the initial configuration $\qstart \in \CF$ towards a configuration
$\qgoal \in \CF$ such that the distance between the gripper and the nearest target pose $T_g$ for some $g=(T_g,s_g), g \in \G$ is minimized.

\section{Framework description}
\label{sec:framework}
First, we give an overview of the components within our framework.
Then we will discuss the proposed scenarios, the grasp annotations, and the baseline planners.

\subsection{Overview}

\begin{figure}[thpb]
  \centering
  \includegraphics[width=\columnwidth]{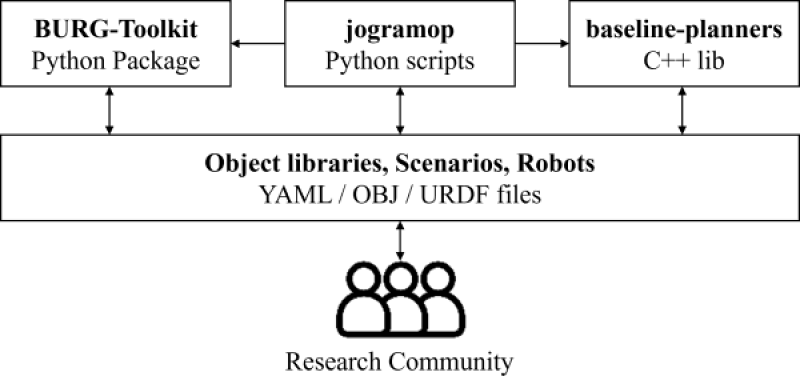}
  \caption{Overview of our framework.
    All code is openly available and all relevant files for the benchmark scenarios are shared with the research community.
    Using our tools, other researchers can contribute their own scenarios in the same format.
}
  \label{fig:framework-overview}
\end{figure}

Fig.~\ref{fig:framework-overview} shows the components of our framework.
The BURG-Toolkit\footnote{
    The BURG-Toolkit is a toolkit for Benchmarking and Understanding Robotic Grasping that we developed.
    A preliminary version of it has been presented at a workshop~\cite{BURG_toolkit_2022} but it has not been previously published.
} 
provides the backbone for describing objects in an object library, composing scenes with these objects, sampling grasp candidates as well as executing them in simulation.
Furthermore, we provide a set of scripts for creating the benchmark scenarios including all relevant annotations and for evaluating and analysing results.
The planners themselves are implemented in C\texttt{++} for best performance.

Using this set of tools, we created 20~fully annotated benchmark scenarios that we share with the research community.
As the tools are openly available, researchers can use them to create their own scenarios, or test their own planners on the provided scenarios.
We are convinced that our philosophy to openly share tools, benchmarks, and methods is fundamental for improving reproducibility in robotics research.

\subsection{Scenarios}

We prepared four different environments: 01 grasping a screwdriver from a shelf, 02 grasping a ball underneath a table, 03 reaching through a narrow gap to grasp a box, and 04 navigating the manipulator through a narrow opening to grasp a banana (see Fig.~\ref{fig:scenarios}).
The difficulties stem from the constricted space in combination with the type of object and its predominant grasp modes.
For example, consider environment~03.
Due to its dimensions, the box can only be grasped from the top or from two of the four sides.
The way we oriented the box behind the walls does not allow easy approach to the sides and hence encourages grasps from the top.

To more systematically investigate the challenges which those environments pose, we introduced five difficulty levels.
In environments~01 and~02, we gradually move the object further back, so that the arm needs to reach further into the confined space.
Conversely, in environments~03 and~04, we gradually reduce the size of the opening to produce a more constricted space.
This gives us 20~scenarios in total and allows us to investigate the effect of the change in difficulty on the performance of the planners.

\begin{figure}[tb]
	\centering
    \vspacebeforefigure
    \begin{subfigure}[t]{0.48\columnwidth}
       \centering
       \includegraphics[width=\textwidth]{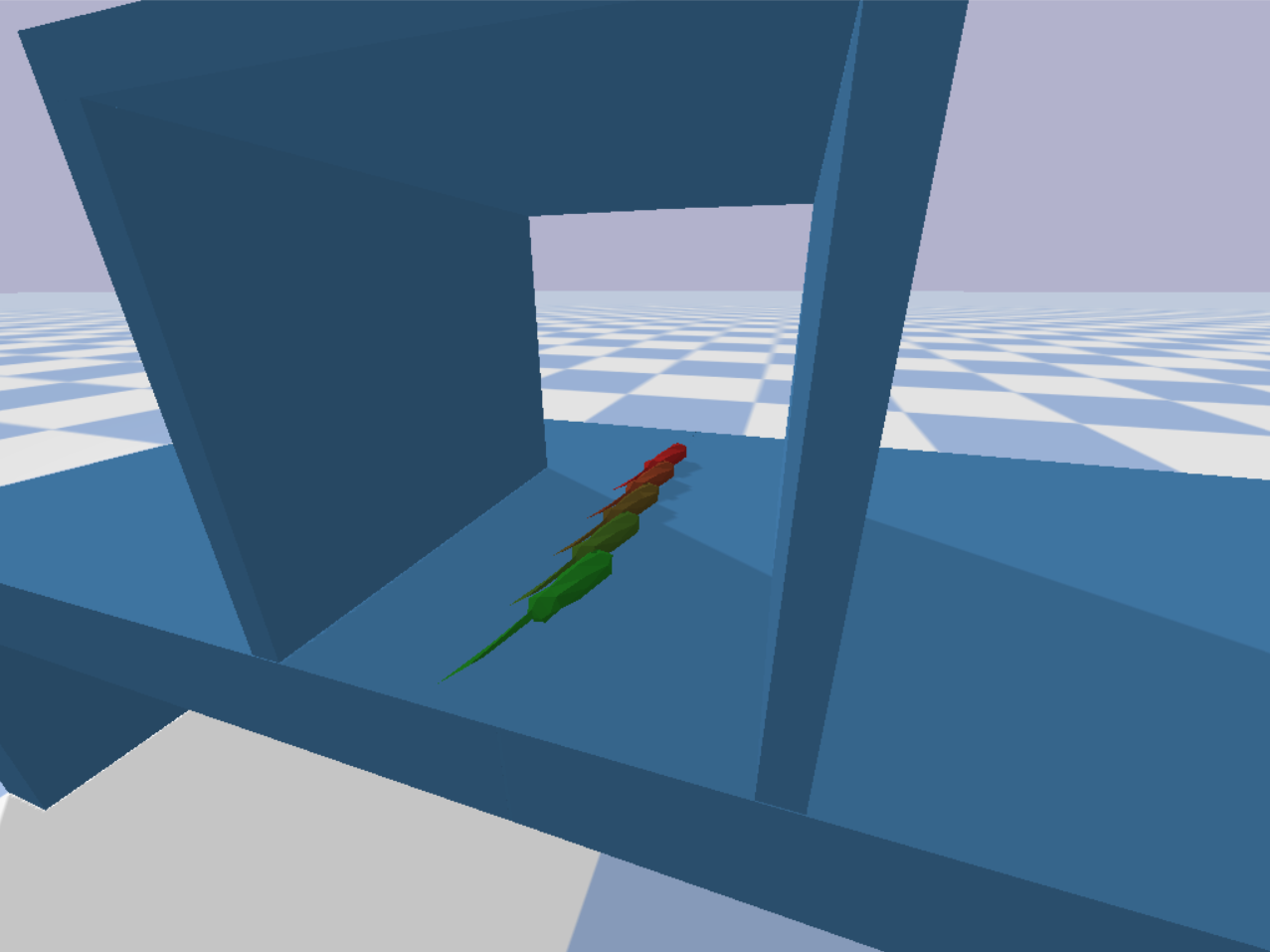}
       \vspacebeforesubcaption
       \caption{Environment 01}
  	\end{subfigure}
  	\begin{subfigure}[t]{0.48\columnwidth}
       \centering
       \includegraphics[width=\textwidth]{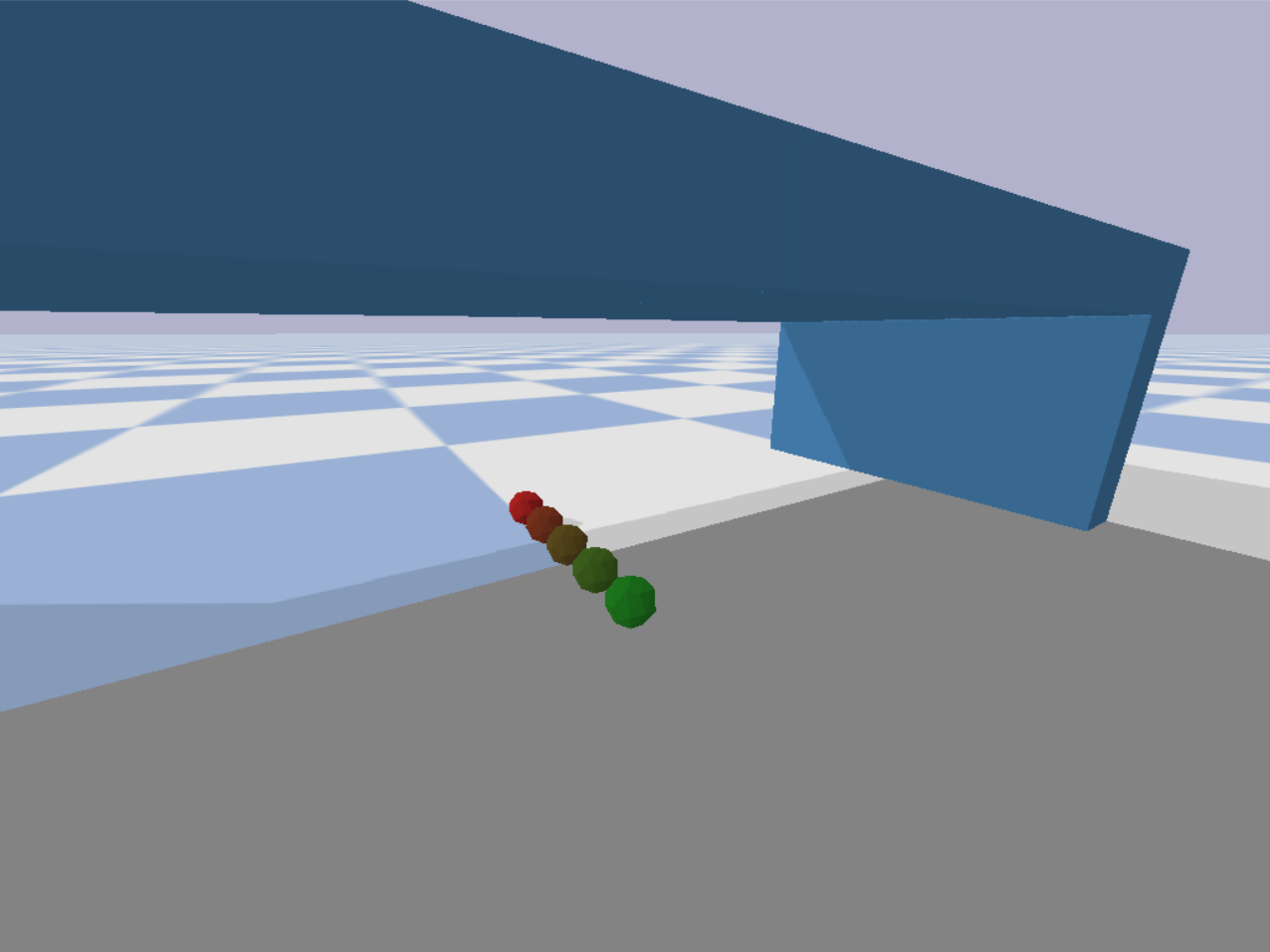}
       \vspacebeforesubcaption
       \caption{Environment 02}
  	\end{subfigure}
  	
  	\vspacebetweensubfigures
  	\begin{subfigure}[t]{0.48\columnwidth}
       \centering
       \includegraphics[width=\textwidth]{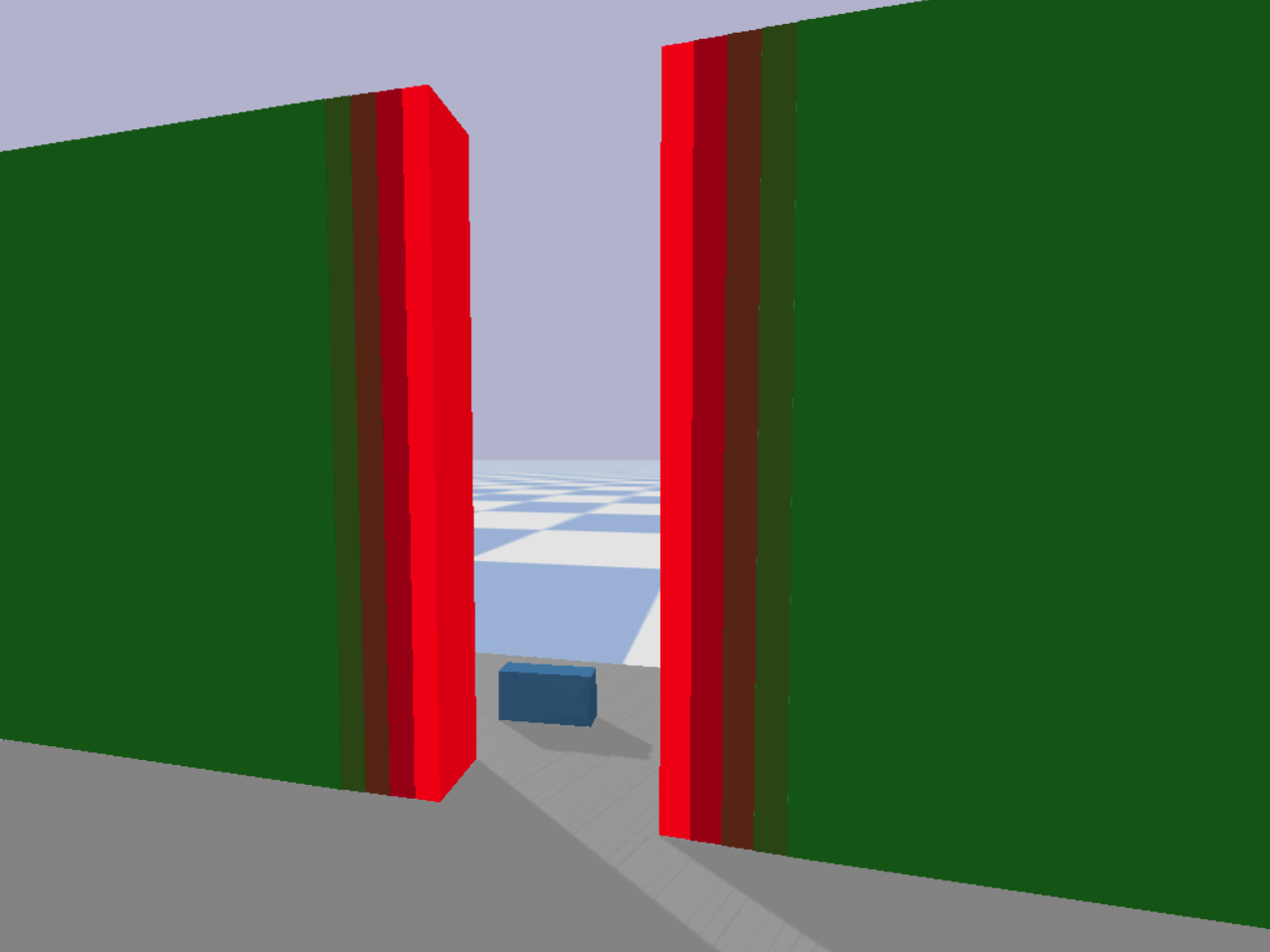}
       \vspacebeforesubcaption
       \caption{Environment 03}
  	\end{subfigure}
  	\begin{subfigure}[t]{0.48\columnwidth}
       \centering
       \includegraphics[width=\textwidth]{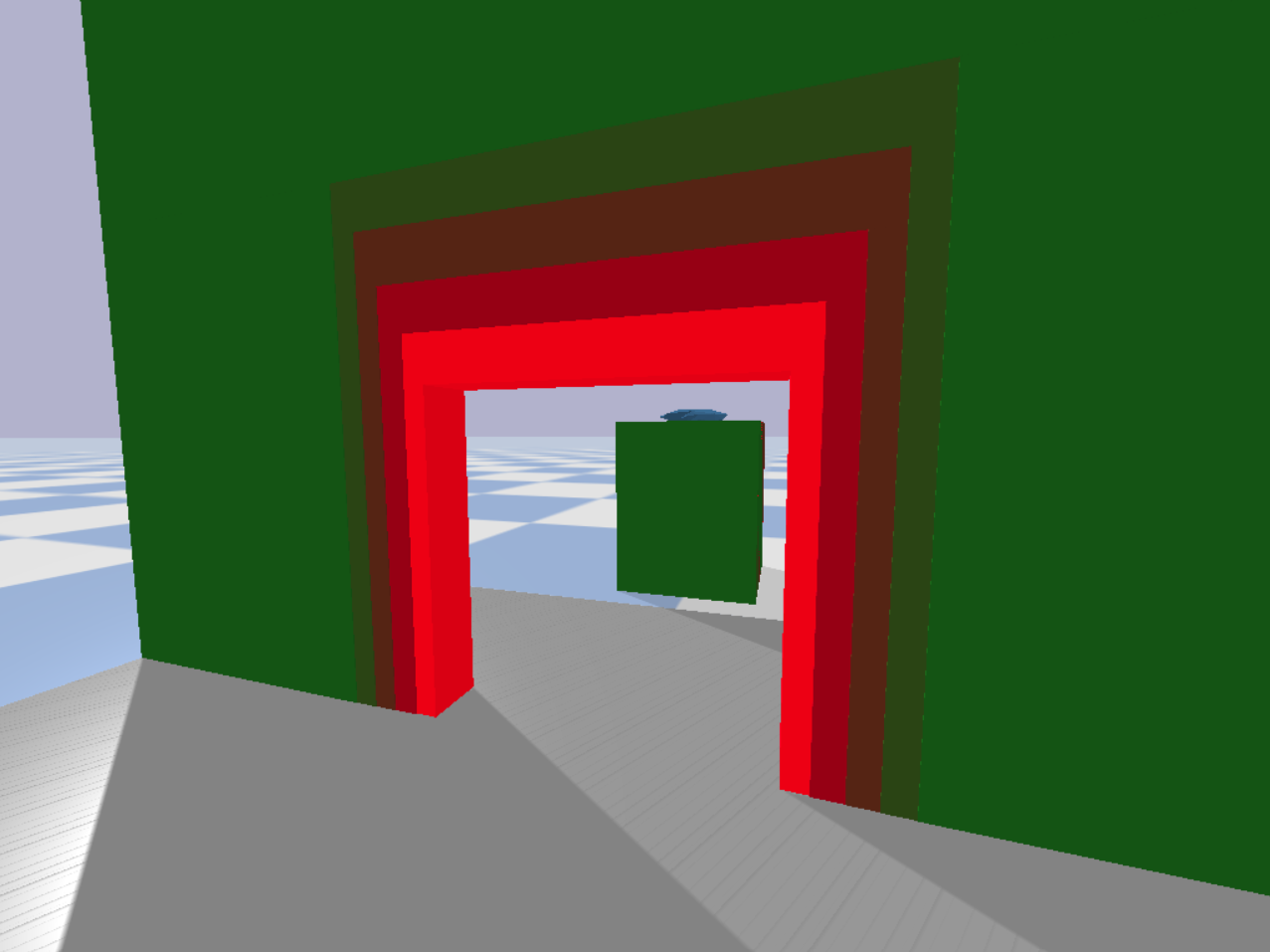}
       \vspacebeforesubcaption
       \caption{Environment 04}
  	\end{subfigure}
  	\caption{Visualization of the difficulty levels in each environment.
        In environments 01 and 02, the object moves further into the confined space, while in environments 03 and 04 the constriction gets narrower.
        }
  	\vspaceafterfigure
  	\label{fig:scenario-difficulties}
\end{figure}

In all scenarios, we use a Franka Panda robot with a two-finger parallel gripper.
The robot is mounted on a platform that mimics a simple holonomic mobile base.
It can move in x and y direction with certain limits (see grey area in Fig.~\ref{fig:scenarios}) but its orientation is fixed.
The base has a dimension of $0.4m \times 0.4m$, i.e. it is small enough for the robot to travel through the opening in environment~04, but in environment~03, it depends on the difficulty level.
In total, the robot has 9~DoF.
It is initialised in the same starting configuration in all scenarios.
Although we only considered this type of robot in this paper, different robots can easily be integrated into the framework.

The target objects are taken from the YCB~benchmark object set~\cite{YCB_Calli2015} which is frequently used for grasping.
They are placed in stable poses on a resting surface.
The obstacles have been created from geometric primitives (e.g. boxes, planes, etc.).
For all objects, including the robot model, we also provide simplified meshes to allow for faster collision checking and distance calculations.

Single-stage approaches for joint grasp and motion planning often make strong assumptions about the objects to grasp, which allow them to use simplified grasp representations (e.g.~\cite{huhConstrainedSamplingBasedPlanning2018}).
By using realistic and diverse object models in our framework, we encourage approaches to directly address this complexity.

\subsection{Grasp Annotations}

In contrast, multi-stage approaches rely on a given set of grasps candidates.
Although the object models can be used to generate new grasp candidates, we provide grasp annotations for all our scenarios.
This makes it easier to get started and allows for a fair comparison.

For each scenario, we provide a set of grasp annotations that is structurally similar to what contemporary grasp synthesis methods produce, i.e., a set of 6-DoF grasp poses along with a stability score.
We use the mesh models of the objects to sample realistic grasp candidates.
Specifically, we use an antipodal-based grasp sampling scheme as these have been shown to be most effective at capturing large parts of the grasp subspace~\cite{Eppner2019_billion_grasps}.
A contact point pair is antipodal (and the grasp force-closure~\cite{Nguyen1988}), if both contact points lie within each other's friction cones.
We achieve this by randomly picking a point from the object's surface as candidate for the first contact point.
Then, we shoot rays within its friction cone and find where they intersect with the mesh to determine candidates for the second contact point.
If the candidate's friction cone encloses the first contact point, we have an antipodal contact point pair.
This defines the grasp position as well as the grasp axis (cf. Fig.~\ref{fig:gripper-frame}).
Finally, we evenly sample a number of approach directions to construct the full 6-DoF pose~$T_g$.

As score, we use a metric similar to~\cite{Fang2020}, which quantifies the amount of friction required for the grasp to be force closure.
The scores are $s_g \in [0, 1]$ with $1$ being the most stable grasp, i.e., requiring the least friction.

After sampling a large amount of grasp candidates, we use a simplified gripper model to filter out colliding grasps.
Finally, we subsample the remaining set to provide 200~grasp candidates for each scenario.
Note that even though the orientation of the objects are identical in the variations of each environment, the grasp annotations vary and are unique for each scenario.

In addition to the grasp set $\G$, with individual grasps $g=(T_g, s_g)$, we also provide a set of inverse kinematics (IK) solutions $\GIK$, with
$\GIK = \{ (q_g',s_g) | q_g' = IK(T_g), q_g' \in \CF, g \in G$\}.
This is to facilitate the use of planners that plan towards a target configuration rather than a task-space goal.
We use a damped least squares method with null-space control to encourage a configuration that is within the joint limits and similar to the starting configuration.
However, the algorithm does not consider collisions and hence only a small number of IK solutions are valid (i.e., collision-free), therefore $|\GIK| \leq |\G|$, and in some scenarios $|\GIK|=0$.
The number of IK solutions for each scenario are listed in Table~\ref{tab:maintab}.

\subsection{Baseline Planners}

Within our framework we implemented two sampling-based planners that adhere to the multi-stage paradigm of joint grasp and motion planning, i.e., both rely on a given grasp set $\G$.
Our first baseline, \JRRT~\cite{vahrenkamp2009humanoid}, is a state-of-the-art planner that jointly plans towards all task-space targets in~$\G$. 
Our second baseline, \IKRRT, is a naive RRT that expands towards a single target in the configuration space, i.e., it plans a separate path for each target in~$\GIK$.

\IKRRT\ is a bidirectional variation of RRT.
One tree is grown from the start configuration and the second one from the target configuration.
This is only possible when a target configuration is known, therefore \IKRRT\ assumes targets from $\GIK$. 
In every step, \IKRRT\  extends both trees towards a randomly chosen configuration $q_r$ until it reaches that configuration or until collision. 
The growth is terminated if both trees approach each other to a distance less than $\gdist$.

The \JRRT\  planner builds a single tree from the start configuration $\qstart$ and attempts to reach any of the target grasp poses in $\G$.
In each step, the tree is expanded towards a random configuration $\qrand \in \C$ (with probability $1-\gb$), or to a randomly chosen grasp target $T_g, g \in \G$ (with probability $\gb$).
The expansion towards a task-space goal is achieved using the Jacobian pseudo-inverse $J^+$~\cite{shkolnikPathPlanning10002009}.
The growth of \JRRT\  is terminated when any of its nodes are within $\gdist$ of the target poses $T_g, g \in \G$.
This planner does not require IK solutions and is able to exploit the null-space of a grasp target, which is particularly useful when dealing with a highly redundant mobile manipulator.
Furthermore, the planner can take into account a given grasp score by biasing the goal sampling towards higher-scoring grasps.

For both variants, we limit the maximum displacement of each joint in all expansion steps to~$\varepsilon$.
The distance $\gdist$ measures the distance between the end-effector and a target grasp and is computed as weighted sum of 3D translation distance plus 3D orientation distance. 
The weight between the translation and rotation distance is set such that 1~mm in the translation distance equals to $1^{\circ}$ of rotational distance.
We therefore report $\gdist$ as a unitless measure.

\begin{figure*}
\centering
\vspacebeforefigure
\begin{subfigure}[t]{0.24\textwidth}
\includegraphics[width=1.0\textwidth]{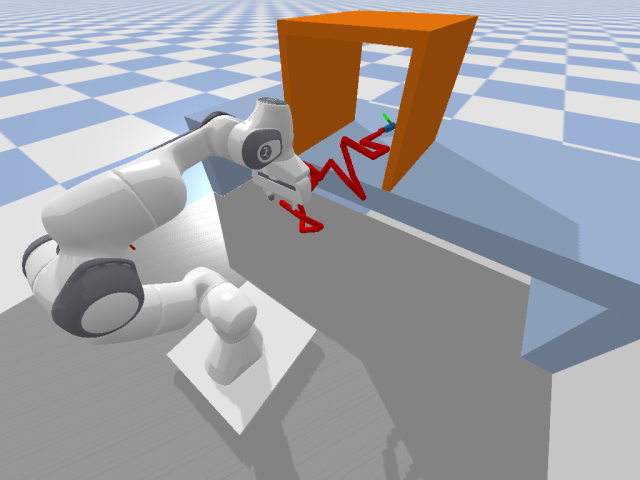}
\vspacebeforesubcaption
\caption{Scenario 014}
\end{subfigure}
\begin{subfigure}[t]{0.24\textwidth}
\includegraphics[width=1.0\textwidth]{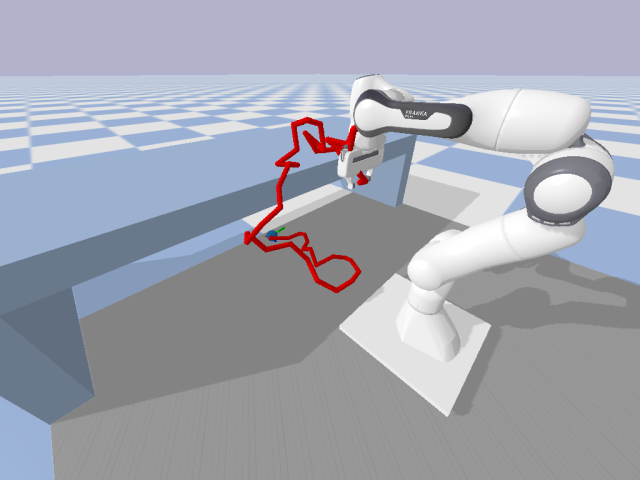}
\vspacebeforesubcaption
\caption{Scenario 022}
\end{subfigure}
\begin{subfigure}[t]{0.24\textwidth}
\includegraphics[width=1.0\textwidth]{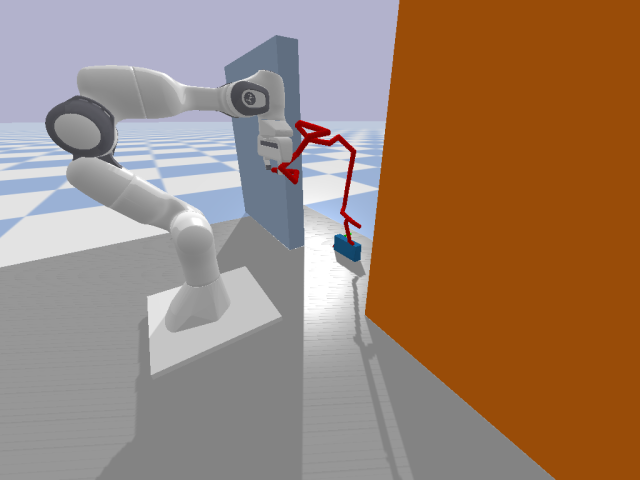}
\vspacebeforesubcaption
\caption{Scenario 031}
\end{subfigure}
\begin{subfigure}[t]{0.24\textwidth}
\includegraphics[width=1.0\textwidth]{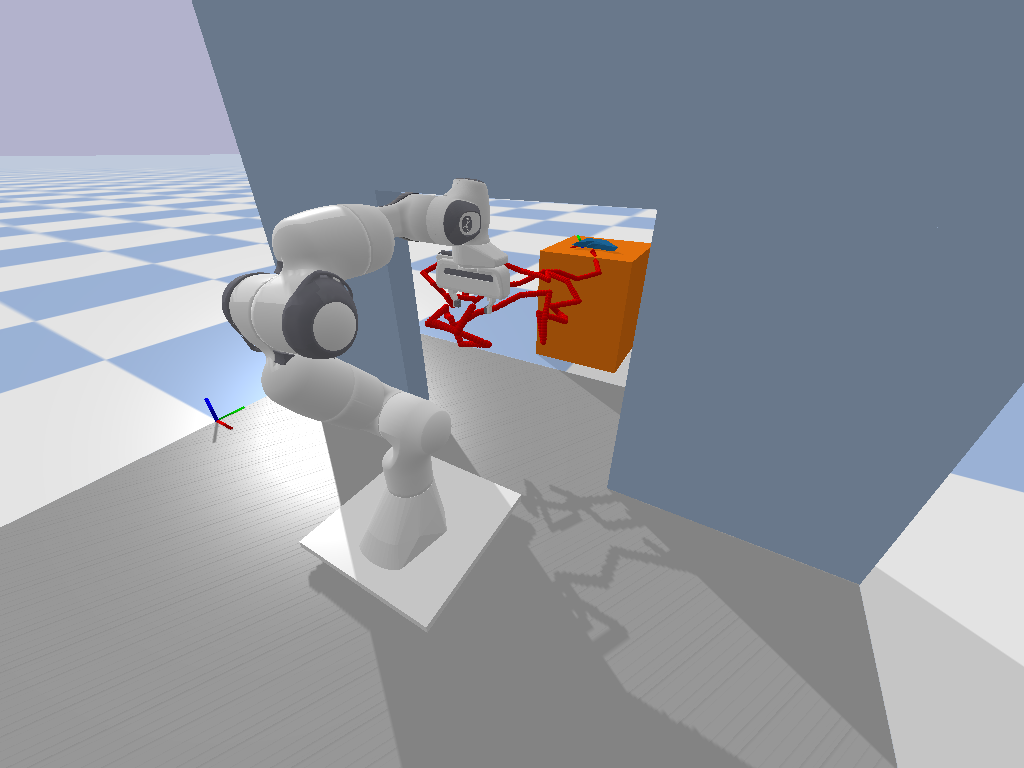}
\vspacebeforesubcaption
\caption{Scenario 042}
\end{subfigure}
\caption{
Examples of trajectories found by the \JRRT\ planner.
\label{fig:trajectories}
}
\end{figure*}

\section{Experiments}
\label{sec:experiments}

We evaluate the performance of both baseline planners on all our scenarios.
As the planners are non-deterministic, we report results over 100~runs.
In each run, the \JRRT\ planner attempts to find a feasible path to any of the grasps in $\G$.
However, the \IKRRT\ baseline can only plan towards a single configuration in $\GIK$ at a time.
Typically, it would sequentially work its way down from the highest-scoring grasp, attempt to find an IK solution, attempt to find a path, and if no path can be found, it would resort to the next-best grasp.
As the IK solutions are already provided within our framework, we simply run the planner 100~times for each of the given targets in~$\GIK$.
Note that calculating the IK solutions for all 200~grasps took between 0.45~s and 0.65~s, which would need to be added to the execution times of \IKRRT.

The planners terminate once the goal is within a distance of less than $\gdist=50$, or if the runtime limit of 120~s is reached.
The goal-bias of both planners was set to $\gb=0.01$ and the resolution to $\varepsilon = 0.1$.

In terms of metrics we report the average success rate, average planning time as well as standard deviation of the planning time.

\begin{table}
\vspacebeforefigure
\caption{
Runtime and success rate of the baseline planners for each of the scenarios.
The planners were run with maximum runtime 120~s.
Time is shown in format mean/stddev (in seconds).
Times for \IKRRT\ do not include calculation of the IK targets.
Note that for scenario 045, we put a maximum runtime of 900~s for \JRRT\ to demonstrate that the scenario is solvable, as no solutions were found in under 120~s.
\label{tab:maintab}
}
\centering
{
\small
\setlength{\tabcolsep}{3pt}
\begin{tabular}{lccll}
\toprule
 \multicolumn{1}{c}{\bf Scenario } & {\bf $|\GIK|$/$*$ } & {\bf $|\G|$} & \multicolumn{1}{c}{{\bf \JRRT }} & \multicolumn{1}{c}{{\bf \IKRRT }} \\ 
\midrule
{\bf 011} & 1/1 & 200  & 91\%: $10.53/33.90$   &  98\%  : 1.52/13.33\\ 
 {\bf 012} & 3/3 & 200  & 95\%: $6.08/26.14$   &  100\%  : 0.02/0.01\\ 
 {\bf 013} & 0/0 & 200  & 95\%: $6.57/26.07$   & --- \\ 
 {\bf 014} & 1/1 & 200  & 97\%: $7.21/20.45$   &  100\%  : 0.04/0.02\\ 
 {\bf 015} & 0/0 & 200  & 80\%: $23.17/24.67$   & --- \\ 
 \midrule
{\bf 021} & 1/1 & 200  & 100\%: $2.63/2.94$   &  100\%  : 0.05/0.03\\ 
 {\bf 022} & 4/4 & 200  & 100\%: $11.26/8.66$   &  100\%  : 0.02/0.01\\ 
 {\bf 023} & 2/2 & 200  & 81\%: $35.13/21.67$   &  98\%  : 1.52/13.33\\ 
 {\bf 024} & 1/1 & 200  & 11\%: $60.00/8.39$   &  100\%  : 0.05/0.04\\ 
 {\bf 025} & 1/1 & 200  & 2\%: $118.56/9.39$   &  11\%  : 108.82/32.07\\ 
 \midrule
{\bf 031} & 10/10 & 200  & 98\%: $1.59/13.32$   &  100\%  : 0.01/0.01\\ 
 {\bf 032} & 11/11 & 200  & 100\%: $0.15/0.29$   &  100\%  : 0.01/0.01\\ 
 {\bf 033} & 15/15 & 200  & 98\%: $1.78/13.31$   &  100\%  : 0.01/0.01\\ 
 {\bf 034} & 15/15 & 200  & 100\%: $0.70/0.98$   &  100\%  : 0.01/0.01\\ 
 {\bf 035} & 8/0 & 200  & 100\%: $1.94/3.05$   &    --- \\ 
 \midrule
{\bf 041} & 0/0 & 200  & 97\%: $4.18/18.76$   & --- \\ 
 {\bf 042} & 5/5 & 200  & 100\%: $4.56/7.24$   &  100\%  : 0.10/0.07\\ 
 {\bf 043} & 2/2 & 200  & 90\%: $25.19/19.80$   &  100\%  : 0.42/0.32\\ 
 {\bf 044} & 5/5 & 200  & 56\%: $85.87/37.95$   &  100\%  : 1.57/1.14\\ 
 {\bf 045} & 4/4 & 200  & 75\%: $604.38/206.88$   &  100\%  : 13.51/8.29\\

\bottomrule
\end{tabular}
}
\end{table}

The quantitative results are shown in Table~\ref{tab:maintab}, where $|\G|$ denotes the number of target grasps and $|\GIK|$ is the number grasps for which collision-free IK solutions were found.
The column $*$ is the number of IK solutions (out of $|\GIK|$) for which \IKRRT\ could find a feasible path at least once.
These results offer interesting insights.
First of all, the naive \IKRRT\ seems to perform well in most scenarios.
However, it becomes very evident how reliant it is on a suitable IK solution.
E.g. in scenarios 013 and 015, no IK solutions are available, hence the method fails entirely while \JRRT\ still reports high success rates.
This demonstrates the need for joint grasp and motion planning approaches.

\begin{figure}
 \centering

 \begin{subfigure}[t]{1.\columnwidth}
 \centering
 \includegraphics[width=1.\textwidth]{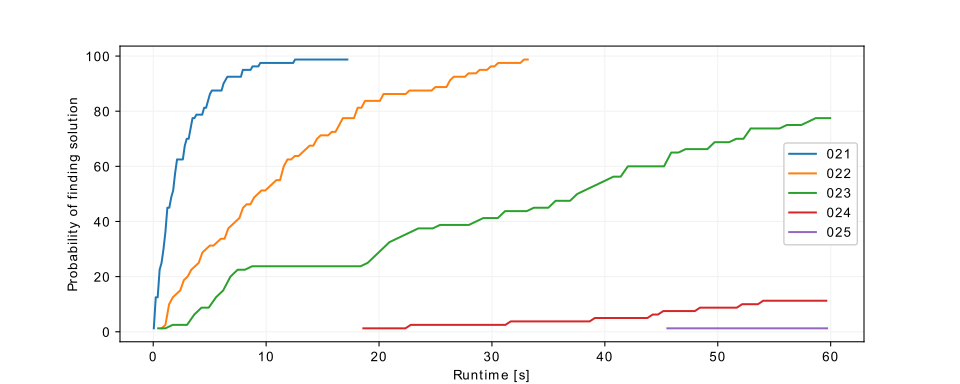}   
 \vspacebeforesubcaption
 \caption{}
 \end{subfigure}
    
\begin{subfigure}[t]{1.\columnwidth}
\centering
\includegraphics[width=1.\textwidth]{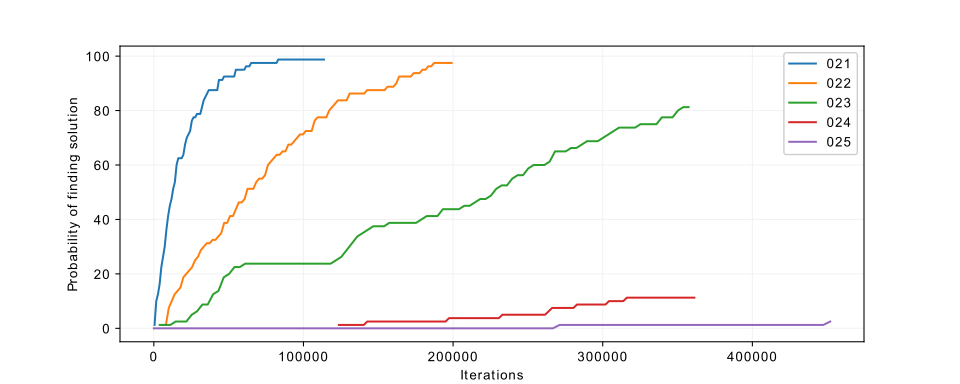}
\vspacebeforesubcaption
\caption{}
\end{subfigure} 
\caption{Success rate curves of \JRRT\ for the five difficulty levels of environment 02 over (a) runtime, and (b) number of random samples.
}
 \label{fig:sr}
\end{figure}

We also see that the difficulty levels of the scenarios have a strong effect on planning performance.
This is most evident in environments~02 and 04.
In environment~02, the robot needs to grasp the ball under the table.
Fig.~\ref{fig:sr} shows the cumulative probability of \JRRT\ finding a solution over its runtime and number of iterations, and it is clearly staggered by difficulty --- the further back the ball is, the longer it takes, and success rates go down.
The difficulty levels have a less clear effect in environments~01 and 03, with environment~03 seemingly being the easiest environment overall.
All of our suggested scenarios have been solved when given sufficient planning time, which shows that they are indeed solvable.

\begin{figure}
\centering
\begin{subfigure}[t]{0.49\columnwidth}
\includegraphics[width=1.0\textwidth]{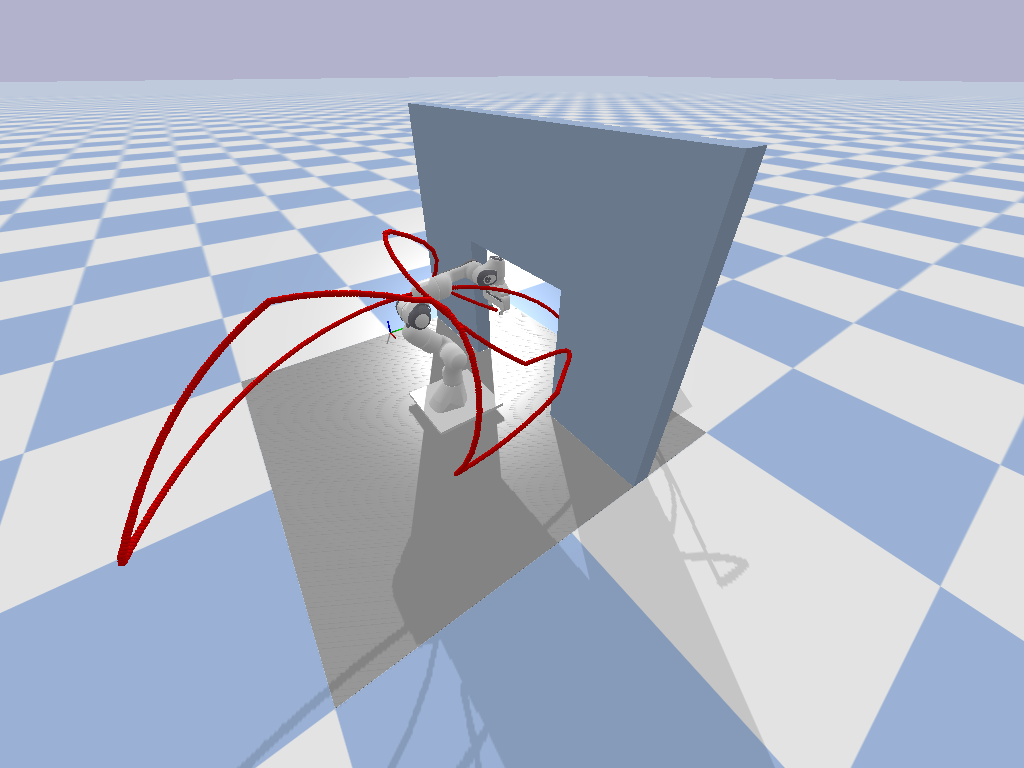}
\vspacebeforesubcaption
\caption{View from room one}
\end{subfigure}
\begin{subfigure}[t]{0.49\columnwidth}
\includegraphics[width=1.0\textwidth]{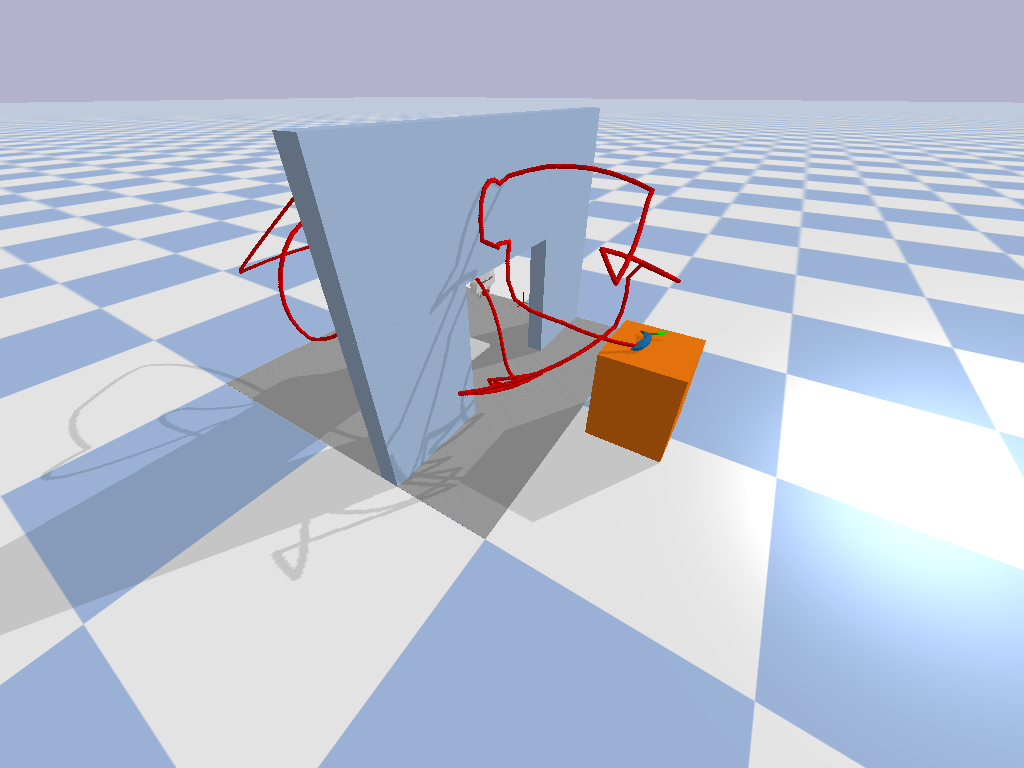}
\vspacebeforesubcaption
\caption{View from room two}
\end{subfigure}

\vspacebetweensubfigures
\begin{subfigure}[t]{1.\columnwidth}
\includegraphics[width=1.0\textwidth]{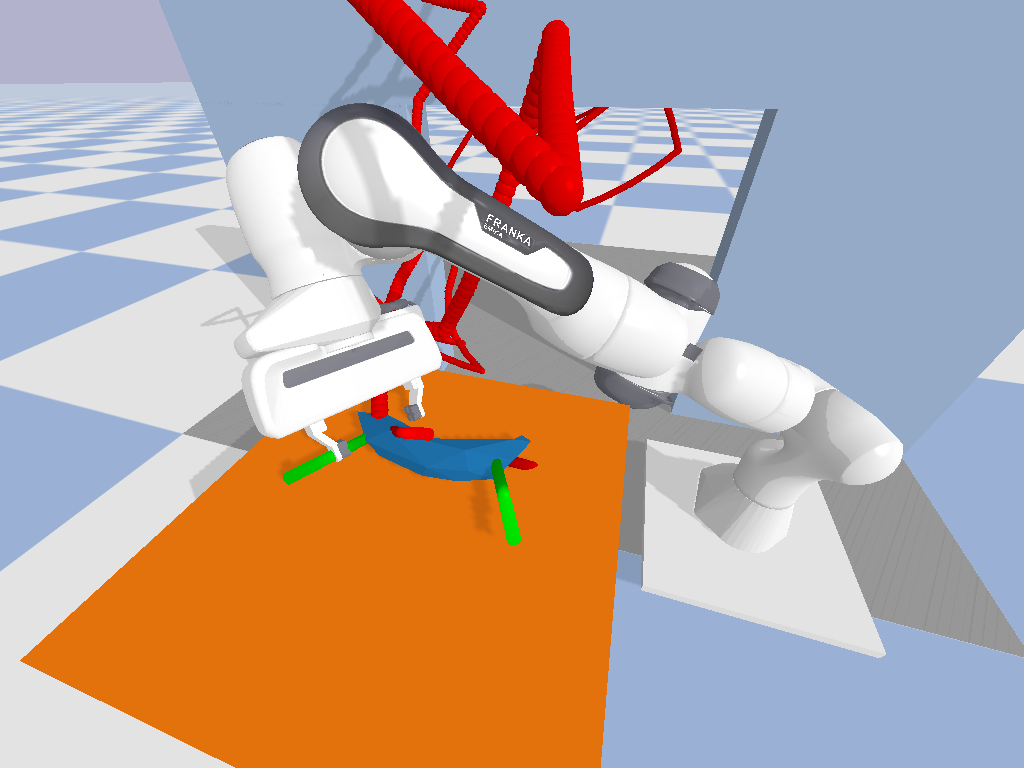}
\vspacebeforesubcaption
\caption{View to final grasping position}
\end{subfigure}

\caption{Example trajectory found by the \IKRRT\ planner for scenario 041.
\label{fig:trajectories041}
}
\end{figure}

Some example trajectories found by the \JRRT\ and \IKRRT\ planner are illustrated in Fig.~\ref{fig:trajectories} and \ref{fig:trajectories041}, respectively.

\section{Conclusions}
\label{sec:conclusions}

In this paper we presented a framework for joint grasp and motion planning in confined spaces.
It includes benchmark scenarios including all relevant annotations as well as baseline planners.
All components are publicly available to the research community and we encourage researchers to use the tools and add their own scenarios and planners.

In our experiments, we demonstrated that the scenarios are indeed challenging and were able to show how differently the two baseline planners \JRRT\ and \IKRRT\ behave.
This suggests that depending on the scenario, a different type of planner might be most suitable.
Further research needs to be done to characterize the difficulties posed by grasping in confined spaces and how they can be tackled by various types of joint grasp and motion planners.



\bibliography{bibliography}

\end{document}